\documentclass[letterpaper, 10pt, conference]{ieeeconf}  % Comment this line out if you need a4paper
\usepackage{amsmath, amssymb}
\usepackage{indentfirst}
\usepackage{graphicx}
\usepackage{subcaption}
\usepackage{caption}
\usepackage{float}
\usepackage{booktabs}
\usepackage{cite}
\usepackage{cleveref}
\Crefname{figure}{Fig.}{Figs.}
\IEEEoverridecommandlockouts                              % This command is only needed if 
\title{\LARGE \bf
Channel-wise Motion Features for Efficient Motion Segmentation
}

\author{Riku Inoue, Masamitsu Tsuchiya, Yuji Yasui
\thanks{The authors are with Honda R$\&$D Co., Ltd., Akasaka, Minato-ku, Tokyo, 107-6238, Japan.
        (Email: \texttt{riku\_inoue@jp.honda})}
}

\usepackage{fancyhdr}

\fancypagestyle{firstpage}{
  \fancyhf{}                      
    
  \fancyhead[C]{%
    \small
    This paper has been accepted to IROS 2024\\
    October 14--18, 2024\\
    Abu Dhabi National Exhibition Centre (ADNEC), Abu Dhabi, UAE
  }
  \fancyfoot[C]{\thepage}     
}

\fancypagestyle{plain}{
  \fancyhf{}              
  
  \fancyfoot[C]{\thepage}  
}

\begin{document}

\maketitle
\thispagestyle{firstpage} 
%\thispagestyle{empty}
%\pagestyle{empty}

%%%%%%%%%%%%%%%%%%%%%%%%%%%%%%%%%%%%%%%%%%%%%%%%%%%%%%%%%%%%%%%%%%%%%%%%%%%%%%%%
\begin{abstract}

For safety-critical robotics applications such as autonomous driving, it is important to detect all required objects accurately in real-time.
Motion segmentation offers a solution by identifying dynamic objects from the scene in a class-agnostic manner.
Recently, various motion segmentation models have been proposed, most of which jointly use subnetworks to estimate Depth, Pose, Optical Flow, and Scene Flow.
As a result, the overall computational cost of the model increases, hindering real-time performance.

In this paper, we propose a novel cost-volume-based motion feature representation, Channel-wise Motion Features.
By extracting depth features of each instance in the feature map and capturing the scene's 3D motion information, it offers enhanced efficiency.
The only subnetwork used to build Channel-wise Motion Features is the Pose Network, and no others are required.
Our method not only achieves about 4 times the FPS of state-of-the-art models in the KITTI Dataset and Cityscapes of the VCAS-Motion Dataset, but also demonstrates equivalent accuracy while reducing the parameters to about 25$\%$.
\end{abstract}

\setlength{\headheight}{4pt}  
\setlength{\headsep}{4pt}  
\pagestyle{plain}
%%%%%%%%%%%%%%%%%%%%%%%%%%%%%%%%%%%%%%%%%%%%%%%%%%%%%%%%%%%%%%%%%%%%%%%%%%%%%%%%

\section{INTRODUCTION}

Understanding the visual scenes is foundational to computer vision in various applications, such as autonomous driving and robotics. 
When these systems operate in real-world environments, they often encounter objects that were not present in the training data. 
A crucial technique for effectively addressing these unknown scenarios is motion segmentation, which identifies moving objects from the scene in a class-agnostic manner \cite{siam2021video}.
This task is a very challenging problem because it requires an understanding of complex information in the scene.
As a result, many recent deep learning-based motion segmentation models showcasing significant advancements employ multiple subnetworks to estimate Optical Flow, Depth, Pose, and Scene Flow, 
in turn raising concerns regarding computational efficiency and speed \cite{neoral2021monocular,yang2021learning,siam2021video,homeyer2023moving,ranjan2019competitive}.

In this paper, we introduce a novel motion segmentation model that achieves real-time performance while maintaining competitive accuracy compared to methods integrating multiple subnetworks. %as illustrated in \Cref{fig:graph}. 
Our approach is inspired by the multi-frame depth estimation model, ManyDepth\cite{watson2021temporal}, and the real-time instance segmentation model, SparseInst\cite{Cheng2022SparseInst}.

ManyDepth converts the 4D Cost Volume calculated from consecutive frames into a 3D Cost Volume by aggregating the channel directions and uses this information to estimate the depth of the target image.
It's observed in the Cost Volume that regions with moving objects show irregular values compared to their surroundings, and ManyDepth introduces a method to address this issue. 
We hypothesize that this characteristic of Cost Volume can be effectively applied to motion segmentation.

SparseInst performs instance segmentation based on instance activation maps that emphasize information-rich regions of each object in a scene.
Each activation map is tailored to highlight a specific instance, subsequently guiding the creation of object masks.
We hypothesized that by aggregating the 3D information from the feature maps produced by the backbone of an instance segmentation model designed to learn instance activation maps, 
we could obtain the 3D motion information of each instance.

\begin{figure}[tp]
    \centering
    \begin{minipage}[t]{.32\hsize}
        \centering
        \includegraphics[width=\linewidth]{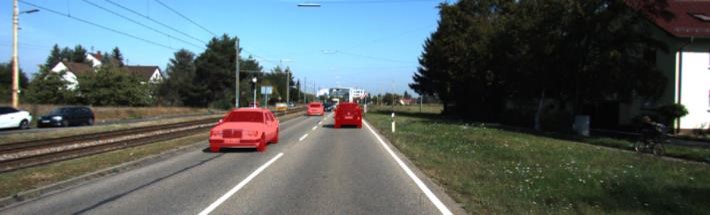}
        %\subcaption{}
        \label{fig:gt1}
    \end{minipage}
    \begin{minipage}[t]{.32\hsize}
        \centering
        \includegraphics[width=\linewidth]{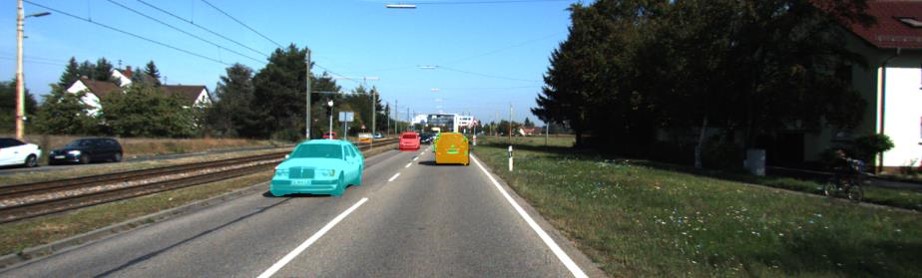}
        %\subcaption{}
        \label{fig:out1}
    \end{minipage} 
    \begin{minipage}[t]{.32\hsize}
      \centering
      \includegraphics[width=\linewidth]{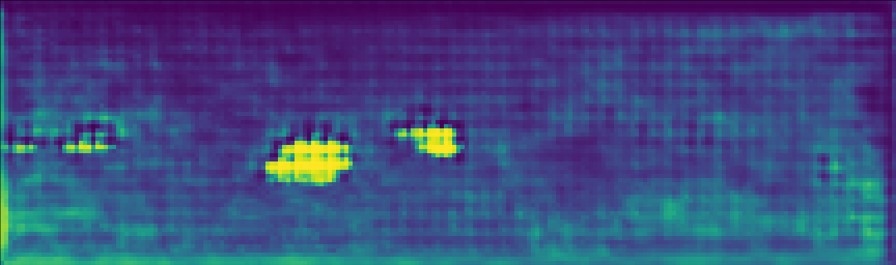}
      %\subcaption{}
      \label{fig:feat1}
    \end{minipage} \\
  
    \centering
    \begin{minipage}[t]{.32\hsize}
        \centering
        \includegraphics[width=\linewidth]{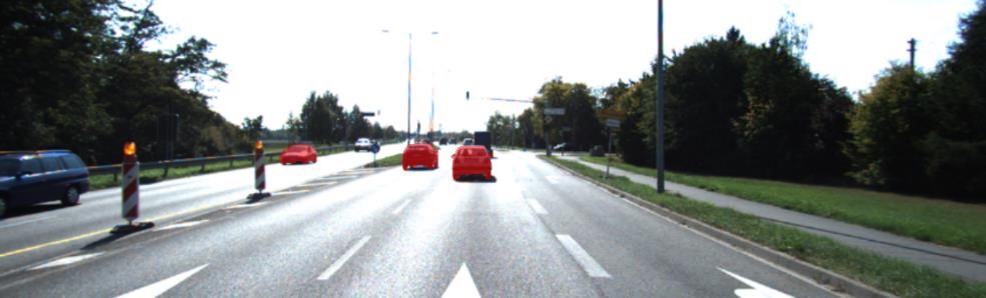}
        %\subcaption{}
        \label{fig:gt2}
    \end{minipage}
    \begin{minipage}[t]{.32\hsize}
        \centering
        \includegraphics[width=\linewidth]{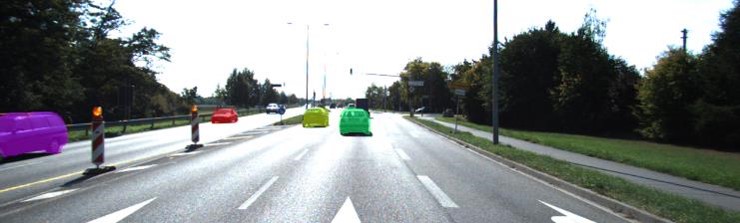}
        %\subcaption{}
        \label{fig:out2}
    \end{minipage}
    \begin{minipage}[t]{.32\hsize}
      \centering
      \includegraphics[width=\linewidth]{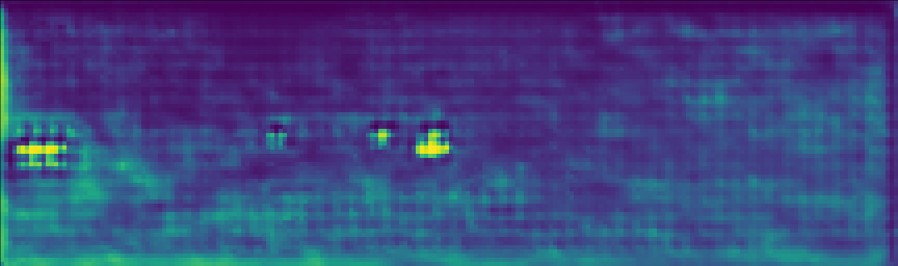}
      %\subcaption{}
      \label{fig:feat2}
    \end{minipage} \\
  
    \centering
    \begin{minipage}[t]{.32\hsize}
        \centering
        \includegraphics[width=\linewidth]{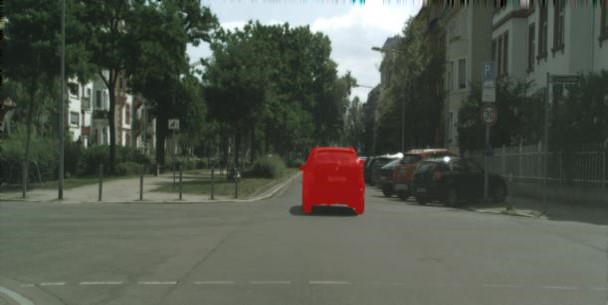}
        %\subcaption{}
        \label{fig:gt3}
    \end{minipage}
    \begin{minipage}[t]{.32\hsize}
        \centering
        \includegraphics[width=\linewidth]{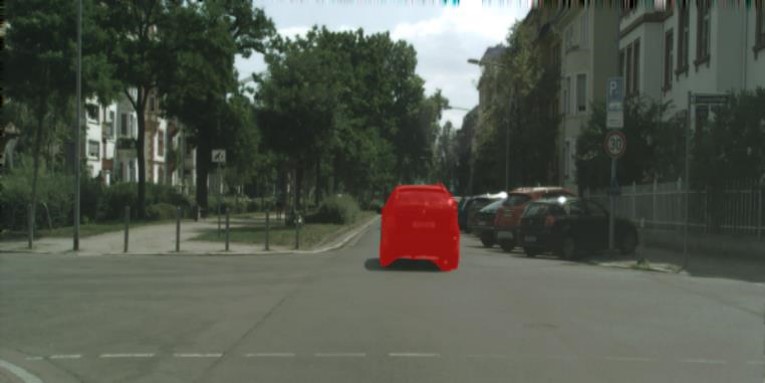}
        %\subcaption{}
        \label{fig:out3}
    \end{minipage}
    \begin{minipage}[t]{.32\hsize}
      \centering
      \includegraphics[width=\linewidth]{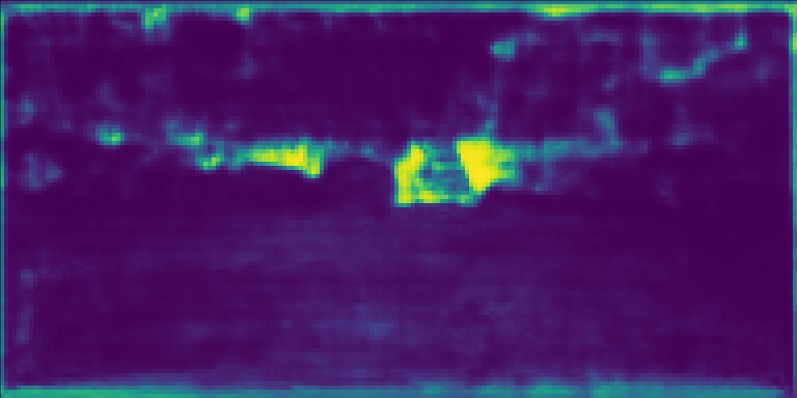}
      %\subcaption{}
      \label{fig:feat3}
    \end{minipage} \\
  
    \centering
    \begin{minipage}[t]{.32\hsize}
        \centering
        \includegraphics[width=\linewidth]{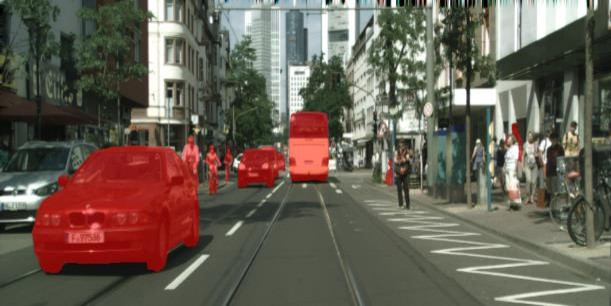}
        \subcaption{Ground Truth}
        \label{fig:gt5}
    \end{minipage}
    \begin{minipage}[t]{.32\hsize}
        \centering
        \includegraphics[width=\linewidth]{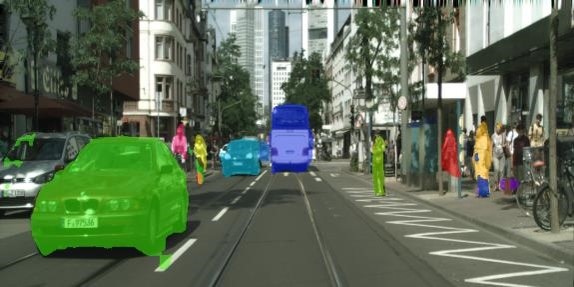}
        \subcaption{Output Mask}
        \label{fig:out5}
    \end{minipage}
    \begin{minipage}[t]{.32\hsize}
      \centering
      \includegraphics[width=\linewidth]{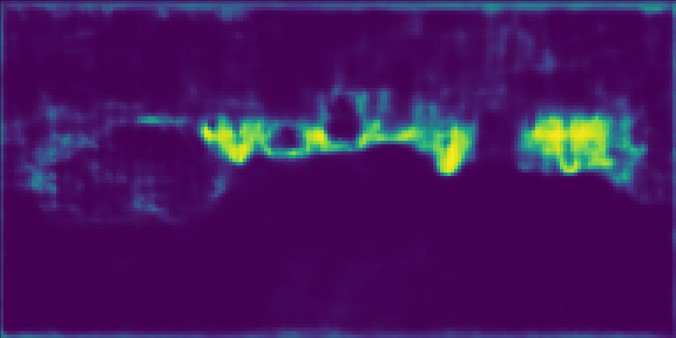}
      \subcaption{Feature Map}
      \label{fig:feat5}
    \end{minipage}
  
    \caption{The proposed model outputs masks and feature maps.
    (a) shows the Ground Truth masks from the VCAS-Motion Dataset\cite{siam2021video}, (b) shows the output motion segmentation masks from the proposed model, and (c) shows the feature maps of Channel-wise Motion Features. 
    The feature maps of Channel-wise Motion Features are described in \cref{Ablation}.}
    \label{fig:masks_and_features}
  \end{figure}

In this paper, we propose Channel-wise Motion Features, a novel representation built on the properties of Cost Volume and instance activation maps.
Channel-wise Motion Features leverages feature maps from a backbone trained similarly to SparseInst to generate instance activation maps and constructs a 4D Cost Volume using a method consistent with ManyDepth, 
aggregating depth-direction information for each channel within the volume.
This represents the 3D motion information of each instance emphasized in each channel.
Moreover, since Cost Volume values for moving objects are inconsistent, this further enhances information beneficial for detecting moving objects.
By integrating a 3D convolutional network designed to produce this representation into a model based on SparseInst, we have achieved an efficient motion segmentation model.
Additionally, when constructing the 4D Cost Volume, we introduce a new depth range setting distinct from ManyDepth, specifically tailored for detecting moving objects.
As shown in \Cref{fig:masks_and_features}, Channel-wise Motion Features show activations for each moving instance, making them effective features for motion segmentation.

In summary, the main contributions of this work are:
\begin{quote}
    \begin{itemize}
    \item We introduce Channel-wise Motion Features, which achieves approximately four times the frames per second and reduces parameters by about 25$\%$, with only a 6.09$\%$ F-measure drop in the KITTI Dataset 
    and a 0.48$\%$ CAQ decrease in the VCAS-Motion Cityscapes Dataset compared to state-of-the-art models.
    \item Through extensive experiments and comparison with other methods, we demonstrate the effectiveness of Channel-wise Motion Features.
     \item We propose a new depth range-setting method for cost volume construction and show its effectiveness through experiments.
    \end{itemize}
\end{quote}

\label{sec:intro}

\section{RELATED WORK}

\subsection{Motion Segmentation}
Motion segmentation, which aims to detect moving objects from images, is recognized as a challenging task, as discussed in various prior works.
The problems include motion degeneracy \cite{yang2021learning}, the need for recognition of motion and 3D geometry \cite{homeyer2023moving}, 
changes in camera ego-motion, lighting changes between consecutive frames, motion blur, and variations in pixel displacement due to different movement speeds \cite{vertens2017smsnet}.
Due to the complexity of motion segmentation, multiple subnetworks have been utilized to infer scene dynamics, 
such as optical flow \cite{tokmakov2019learning}, depth, optical flow\cite{vertens2017smsnet}, depth, ego-motion, optical flow\cite{siam2021video, ranjan2019competitive, homeyer2023moving}, 
depth, optical flow, optical expansion\cite{yang2021learning, neoral2021monocular}, and optical flow combined with LiDAR\cite{rashed2019fusemodnet}.
As a result, the real-time performance of these models is compromised, making their application in real-world systems, such as autonomous driving, challenging. 
On the other hand, our proposed model effectively addresses the trade-off between accuracy and inference speed, demonstrating real-time capabilities that make it viable for real-world applications.

\subsection{Cost Volume}

\begin{figure}[tp]
    \centering 

    \begin{minipage}[t]{.45\hsize}
        \centering
        \vspace{0.25cm} 
        \includegraphics[height=1.3cm, width=3.8cm]{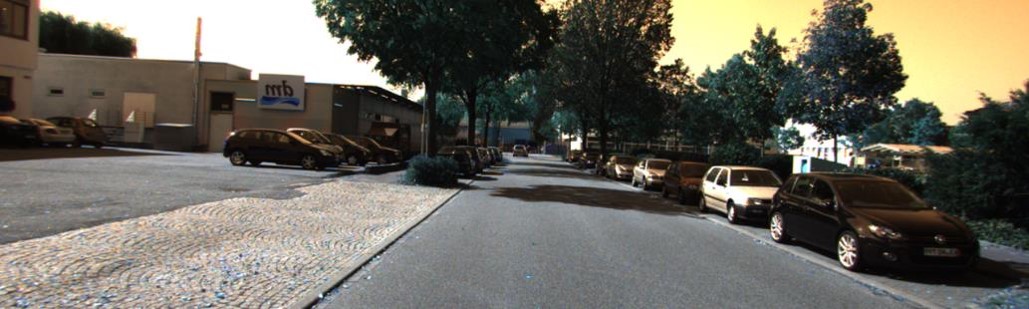}
        \subcaption{Input image of scenes without any moving objects.}
        \label{fig:consistency_rgb}
    \end{minipage} 
    \hfill
    \begin{minipage}[t]{.45\hsize}
        \centering
        \vspace{0.25cm} 
        \includegraphics[height=1.3cm, width=3.8cm]{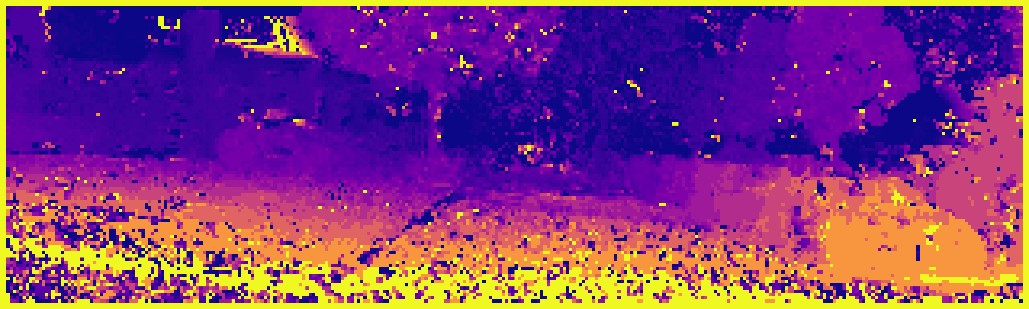}
        \subcaption{Depth from Cost Volume}
        \label{fig:consistency_cost}
    \end{minipage} \\

    \vspace{0.5em} 

    \begin{minipage}[t]{.45\hsize}
        \centering
        \includegraphics[height=1.3cm, width=3.8cm]{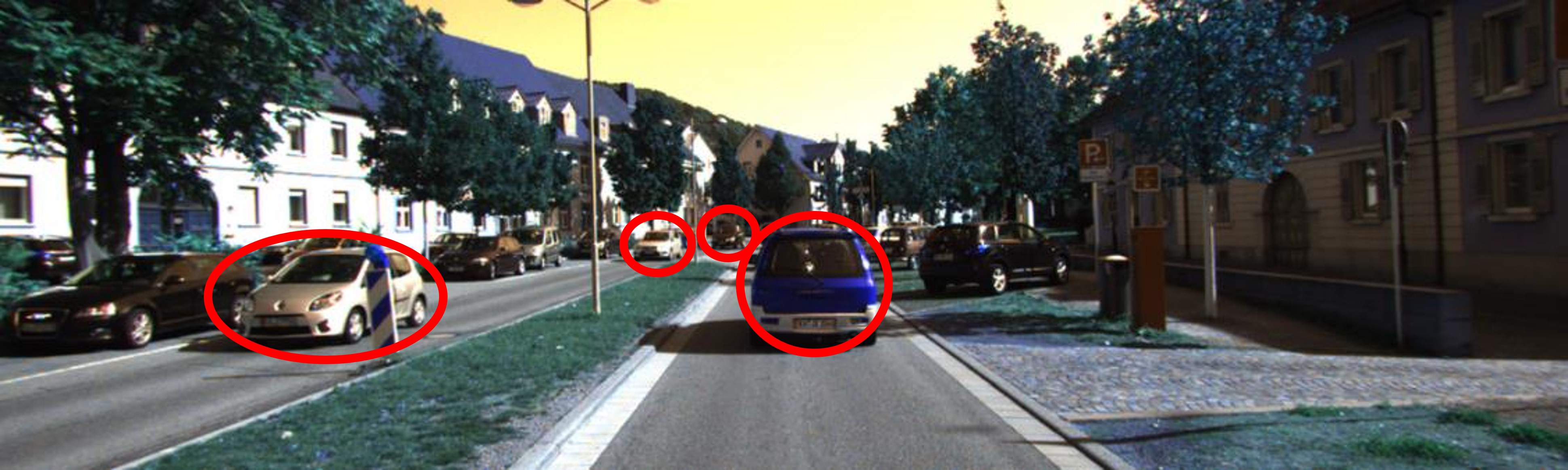}
    \end{minipage} 
    \hfill
    \begin{minipage}[t]{.45\hsize}
        \centering
        \includegraphics[height=1.3cm, width=3.8cm]{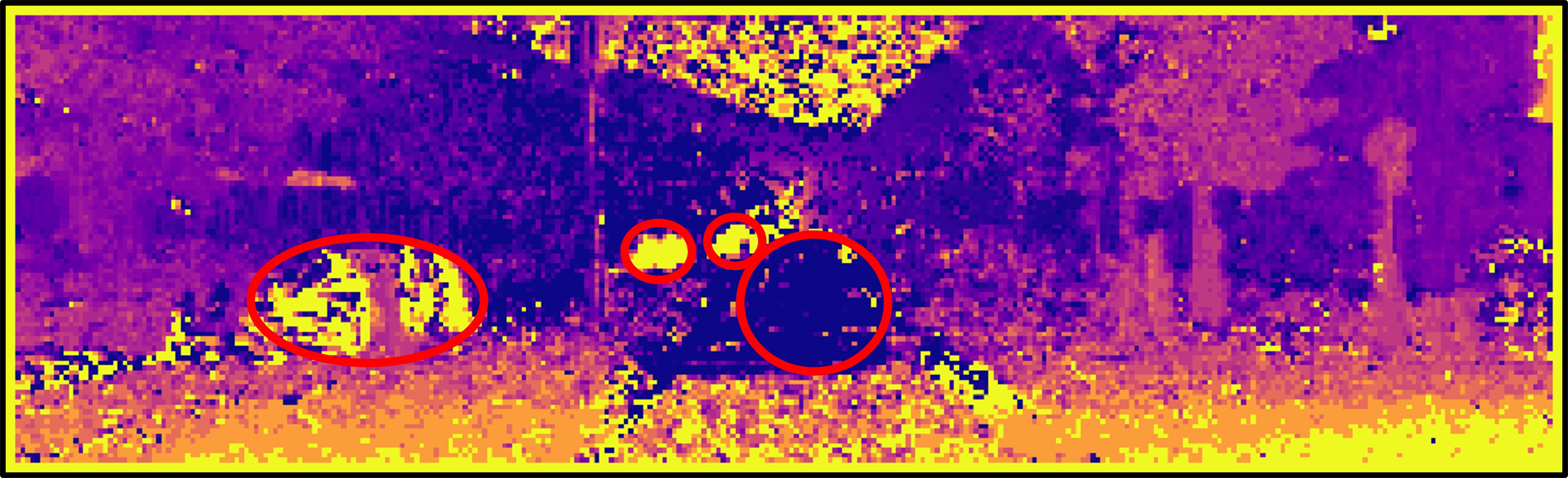}
    \end{minipage} \\

    \vspace{0.2em} 

    \begin{minipage}[t]{.45\hsize}
        \centering
        \includegraphics[height=1.3cm, width=3.8cm]{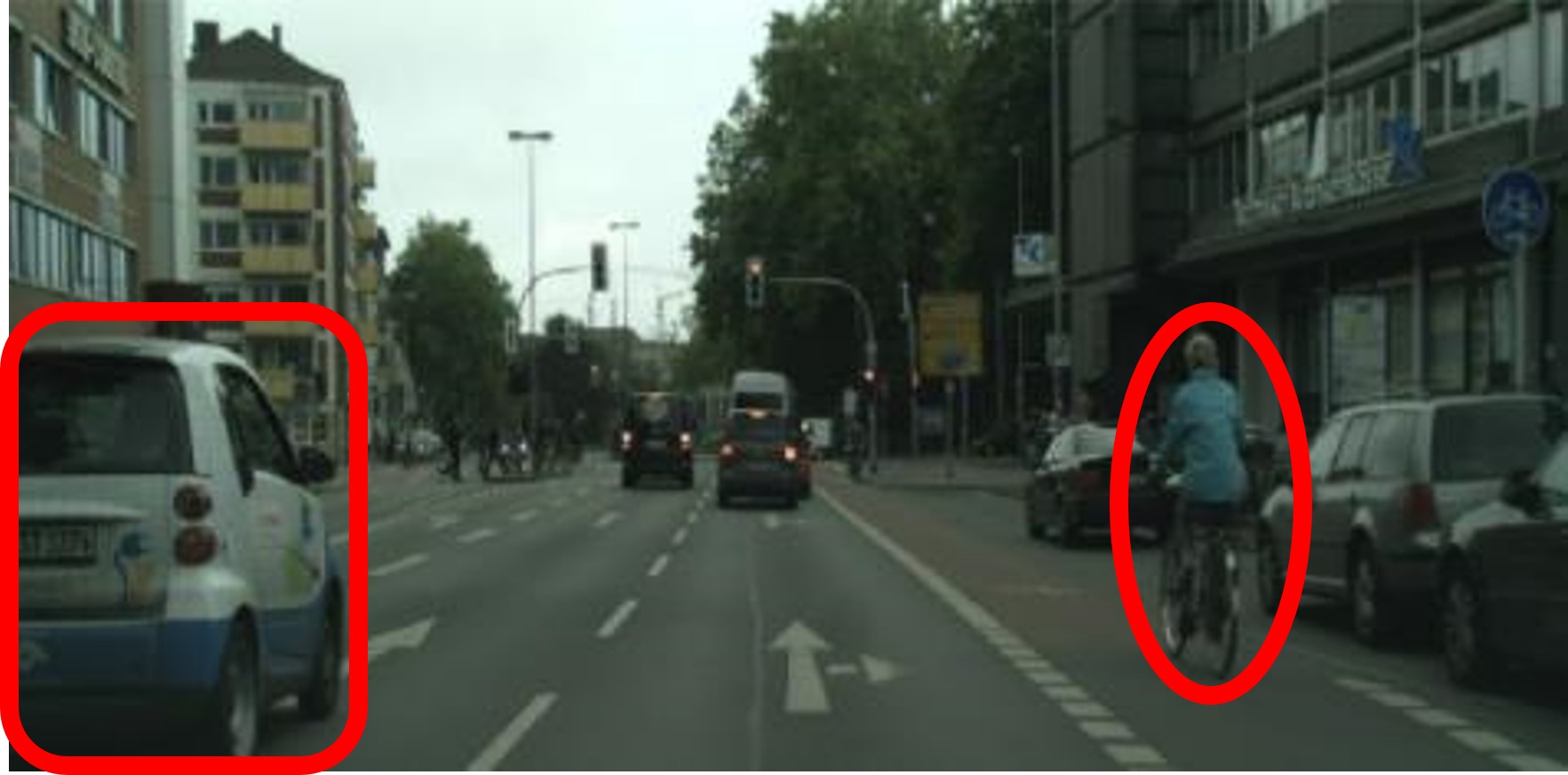}
        \subcaption{Input image of scenes with moving objects present.}
        \label{fig:inconsistency_rgb}
    \end{minipage} 
    \hfill
    \begin{minipage}[t]{.45\hsize}
        \centering
        \hspace*{0.015cm}
        \includegraphics[height=1.3cm, width=3.8cm]{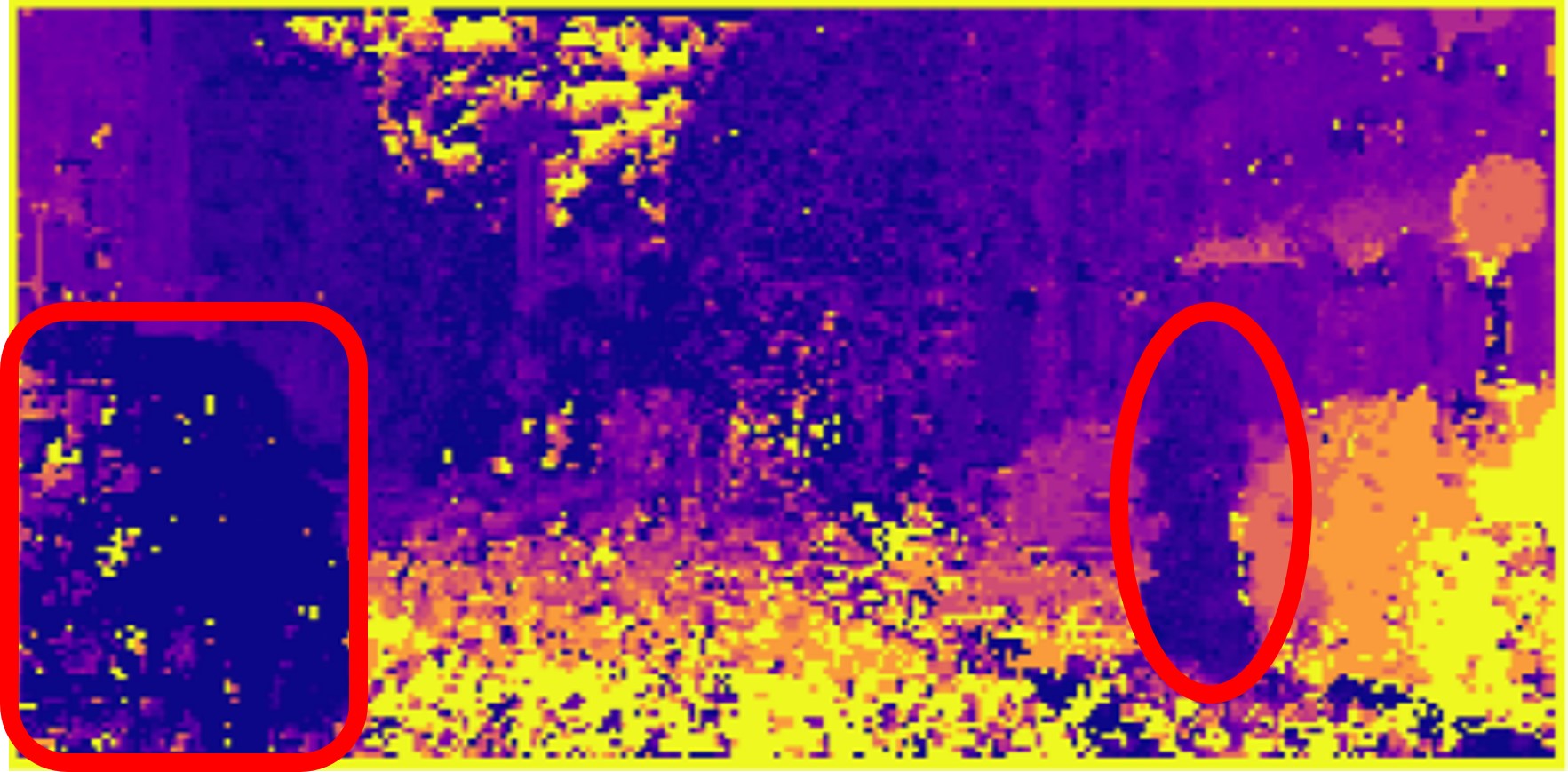}
        \subcaption{Depth from Cost Volume}
        \label{fig:inconsistency_cost}
    \end{minipage}
    \hfill
    \caption{Inconsistency value of cost volume in dynamic object region.
    (a) and (c) are input images without and with moving objects present, respectively. 
    (b) and (d) represent the lowest values extracted along the depth direction from the cost volumes corresponding to their respective input images. 
    In (b), due to the absence of moving objects, it appears continuous, much like a depth map. 
    However, in (d), the moving object indicated by the red circle is clearly discontinuous in value compared to its surroundings.
    }
    \label{fig:inconsistency_cost_value}
\end{figure}

Cost volumes store data matching costs at corresponding pixels between two frames and are commonly used in 
multi-frame depth estimation\cite{watson2021temporal,guizilini2022multi,wimbauer2021monorec}, stereo matching\cite{xu2023iterative,lipson2021raft}, and optical flow estimation\cite{hui2020liteflownet3,teed2020raft}.
In self-supervised multi-frame depth estimation, the source image is warped by considering only ego-motion, which is suffered by moving objects, when acquiring monocular depth information from the cost volume.
ManyDepth\cite{watson2021temporal} addresses this problem through consistency with single-frame depth networks, and DepthFormer\cite{guizilini2022multi} adopting the same approach.
MonoRec\cite{wimbauer2021monorec} handles moving objects using dynamic masks predicted by MaskModule.

Our model exploits the problem of moving objects in Cost Volume to enhance motion segmentation information.
As seen in \Cref{fig:inconsistency_cost_value}, in regions where a moving object is present, the Cost Volume has irregular values relative to the values in the surrounding region.
Channel-wise Motion Features, which was created by taking advantage of this fact is an effective and efficient representation for moving object detection.

\begin{figure*}[tp]
    \centering
    \vspace{0.2cm} 
    \includegraphics[width=0.9\linewidth]{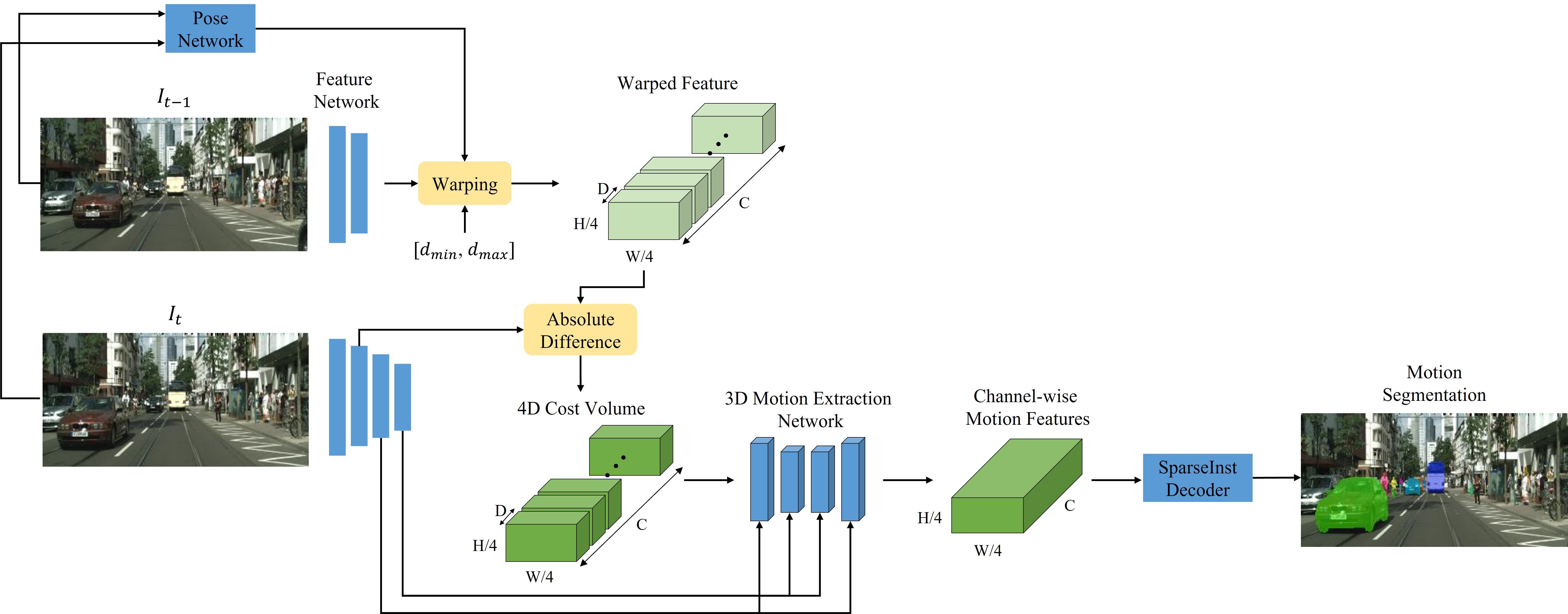}
    \caption{Overview of our proposed model. 
    From two consecutive frames, $I_{t-1}$ and $I_t$, the Pose Network estimates the camera's ego-motion. 
    Then, the Feature Network, which shares weights, extracts features from each of the two frames separately.
    By utilizing this estimated motion along with a predefined set of depth values, we warp the feature map $F_{t-1}$ to align it with the perspective of $I_t$. 
    The 3D Motion Extraction Network then transforms the 4D Cost Volume($D\times C \times H \times W$) into Channel-wise Motion Features($C \times H \times W$), enabling the extraction of motion information for each channel. 
    Finally, leveraging the instance activation maps, the SparseInst Decoder estimates the segmentation masks.}
    \label{fig:architecture}
\end{figure*}

\subsection{Instance Segmentation}
The task of instance segmentation is to segment object instances and estimate the mask and class of each.
As a pioneering model, Mask R-CNN\cite{he2017mask} extends the framework of Faster R-CNN\cite{ren2015faster}, an object detection model, by adding a mask branch for predicting segmentation masks, setting a new standard in instance segmentation.
Region-based methods such as Mask R-CNN first detect object bounding boxes as in Faster R-CNN, and then RoI operations, such as RoI-Pooling\cite{ren2015faster} or RoI-Align\cite{he2017mask} are applied.
These operations extract region features for estimating object classes and segmentation masks.
Subsequent research has primarily focused on addressing the low-quality segmentation masks predicted by Mask R-CNN \cite{kirillov2020pointrend,tang2021look} %\cite{kirillov2020pointrend,cheng2020boundary,yuan2020segfix,tang2021look} 
and improving the precision of bounding box detections \cite{cai2018cascade, chen2019hybrid}.

Another class of methods, instance activation based methods, has been proposed to learn the features of these pixels in order to highlight the information-rich regions of each foreground object.
Typical of such methods are those based on center activation \cite{xie2020polarmask,tian2020conditional}, which represent objects with center pixels instead of bounding boxes and segment them using center features.
SOLO \cite{wang2020solo,wang2020solov2} generates mask kernels for segmentation by object centers. %\cite{xie2020polarmask,tian2020conditional,tian2020fcos}
In recent years, several real-time instance segmentation models of instance activation methods have been proposed \cite{Cheng2022SparseInst, he2023fastinst}.
SparseInst \cite{Cheng2022SparseInst} proposes a sparse set of instance activation maps as a new object representation and achieves fast and high performance.
Our method is based on SparseInst, but the base model alone has only low accuracy in motion segmentation. 
On the other hand, by incorporating our proposed Channel-wise Motion Features into SparseInst, we can improve the performance of motion segmentation.

\label{sec:related}

\section{METHOD}

In this section, we first describe the overall architecture of our proposed model, then discuss each component of detail. 

\subsection{Overall Architecture}
As shown in \Cref{fig:architecture}, our network architecture is built upon the structure of SparseInst\cite{Cheng2022SparseInst}.

Given the consecutive frames $I_{t-1}$, $I_{t}\in\mathbb{R}^{3 \times H \times W}$, we first apply the Pose Network following \cite{watson2021temporal} and predict relative camera pose $T_{t \rightarrow t-1}$.
We apply a Feature Network with shared weights to consecutive frames separately, obtaining a feature map $F^1_{t-1}$ from $I_{t-1}$, and three feature maps $F_{t}^1$, $F_{t}^2$, $F_{t}^3$ from $I_{t}$. 
The Feature Network is composed of the first four layers of ResNet50\cite{he2016deep}, and the resolution of $F^1_{t}$, $F^2_{t}$, and $F^3_{t}$ are $1/4, 1/8$, and $1/16$ of the input image, respectively.
The warped feature $F_{t-1 \rightarrow t}^1\in\mathbb{R}^{D \times C \times H \times W}$ is obtained by warping $F_{t-1}^1$ to the viewpoint of $I_{t}$ using $T_{t \rightarrow t-1}$, 
with the depth range represented by $d_{min}$ and $d_{max}$ which are manually set.
The methodology for determining $d_{min}$, $d_{max}$ is explained in Section \ref{depth range}.
We then build a 4D Cost Volume calculating the absolute difference between the $F_{t-1 \rightarrow t}^1$ and $F^1_{t}$.
Channel-wise Motion Features, which aggregate 3D information for each channel, are generated by passing the 4D Cost Volume and $F_{t}^2$, $F_{t}^3$ through a 3D Motion Extraction Network.
In the end, we estimate the instance motion segmentation mask by feeding the Channel-wise Motion Features into an IAM-based Segmentation Decoder\cite{Cheng2022SparseInst}.

\subsection{Channel-wise Motion Features}
Recently proposed instance segmentation models have been successful in using the activation of feature maps\cite{Cheng2022SparseInst, he2023fastinst}.
Similarly, in our model, we believe that the Feature Network can extract features regarding activation for each instance in the input image.
A key idea in our method is computing and aggregating three-dimensional matching between instance activation features of different frames obtained from the Feature Network, enabling the extraction of motion information of instances.
Below we provide details of our proposed method: Channel-wise Motion Features.

\subsubsection{Construction of the 4D Cost Volume}

Cost volume captures the correspondence between pixels in the different views. 
Our method constructs a 4D cost volume following the multi-frame depth estimation technique\cite{watson2021temporal}.

In this setup, the $I_{t-1}$ serves as the source image, and $I_{t}$ as the target image.
The relative camera pose between these two frames is determined through the Pose Network. 
We generate the warped feature volume $F_{t-1 \rightarrow t}\in\mathbb{R}^{D \times C \times H \times W}$ by warping the source features $F_{t-1}$ to the viewpoint of the target image $I_{t}$ 
using a plane of depth value $d_i$, linearly sampled within the depth range from $d_{min}$ to $d_{max}$ along with the estimated relative camera pose.
The warped feature volume is computed as follows:
\begin{equation}
    F_{t-1 \rightarrow t}(d_i) = F_{t-1} \langle proj (d_i,T_{t \rightarrow t-1},K) \rangle
\end{equation}

where $proj()$ denotes projection function provides the 2D coordinates of the sampled depth value $d_i$ in $F_{t-1}$ and $\langle \rangle$ is the sampling operator.
$K\in\mathbb{R}^{3 \times 3}$ denotes a camera intrinsics.

Subsequently, the element of the 4D Cost Volume \( 4DCV \) at depth \( d_i \) is given by:
\footnotesize 
\begin{equation}
    \begin{split}
    4DCV(d_i, c, x, y) = 
    | & F_{t-1 \rightarrow t}(d_i, c, x, y) - F_{t}(c, x, y) |
    \end{split}
\end{equation}
\normalsize 
where $| |$ denotes the absolute difference, \( x \) and \( y \) are spatial coordinates, and \( c \) represents the channel dimension.
This computation is performed for each depth, spatial location, and channel, yielding a 4D representation.

\subsubsection{Channel-wise Motion Features Extraction}

We present a detailed methodology for transforming the 4D Cost Volume into Channel-wise Motion Features.
Channel-wise Motion Features aggregates 3D information per channel and computes motion information between consecutive frames.
In multi-frame depth estimation \cite{wimbauer2021monorec, watson2021temporal}, the 4D Cost Volume is aggregated along the channel direction to construct a 3D Cost Volume with $D \times H \times W$. 
On the other hand, our approach aggregates information along the depth direction, resulting in a three-dimensional representation of $C \times H \times W$.

\begin{figure}[tp]
    \centering
    \vspace{0.25cm} 
    \includegraphics[width=0.8\linewidth]{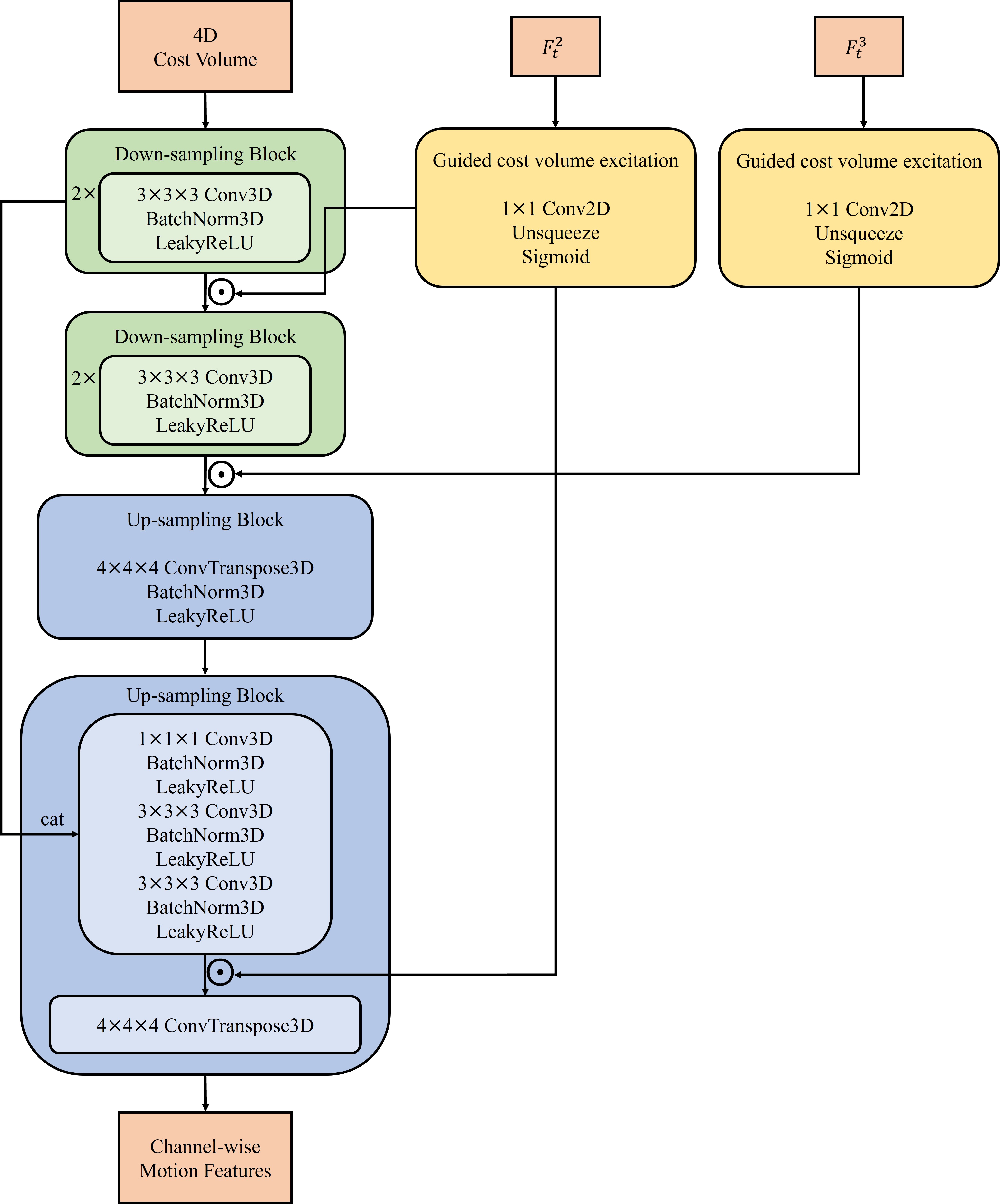}
    \caption{3D Motion Extraction Network.
    \textit{Cat} denotes a concatenation operation.
    The multiplication of numbers described before convolution represents the size of the convolution kernel. 
    In the $1 \times 1 \times 1$ 3D convolution of the second down-sampling block, we adjust the dimensions in the depth direction.}
    \label{fig:3dregular}
\end{figure}

For the transformation of the 4D Cost Volume, we employ a network similar to the 3D Regularization Network presented in \cite{xu2023iterative}.
We made the 3D Regularization Network more lightweight for computational efficiency and repurposed it as a feature extractor  for motion detection. 
Throughout this paper, we will refer to our adapted version of the 3D Regularization Network as the 3D Motion Extraction Network. 
The architecture of the 3D Motion Extraction Network is illustrated in \Cref{fig:3dregular}.
The 3D Motion Extraction Network comprises two down-sampling blocks and two up-sampling blocks.
Each down-sampling block contains two 3D convolutions of size $3 \times 3 \times 3$.
On the other hand, the up-sampling block consists of a block designed with a single $4 \times 4 \times 4$ 3D transposed convolution, 
and a block designed with a $4 \times 4 \times 4$ 3D transposed convolution and two $3 \times 3 \times 3$ 3D convolutions.
The 3D Regularization Network adopts guided cost volume excitation\cite{bangunharcana2021correlate}, which enhances geometric features by applying spatially varying updates to the cost volume feature map.
Transform the feature map, which matches the scale of the cost volume, to align with its channel dimensions and employ it as weights. 
Utilizing these feature map weights, the guided cost volume excitation at scale $i$ can be represented as follows:
\begin{equation}
w = \sigma(f(F_i)) \\
\end{equation}

\begin{equation}
C_i^{'} = w \odot C_i
\end{equation}

where $f$ denotes 2D point-wise convolution, $\sigma$ is the sigmoid function, and $\odot$ is the Hadamard product.
In the final layer of the 3D Motion Extraction Network, 3D transposed convolution is applied to reduce the dimension in the $D$ direction to 1, 
we obtain Channel-wise Motion Features of size $C \times H \times W$. 

\subsection{Setting Depth Range} \label{depth range}

When constructing the Cost Volume, selecting the optimal depth range $[d_{min}, d_{max}]$ is important. 
If the chosen range is excessively broad beyond what is necessary for depth estimation, it leads to sparse depth value sampling, compromising the accuracy of dense matching. 
Conversely, increasing the depth samples for detailed matching amplifies computational demands. 
For efficient dense matching with minimal samples and computational cost, choosing the right depth range is essential.

\begin{figure}[tp]
    \centering
    \vspace{0.2cm} 
    \begin{minipage}[t]{.47\linewidth}
        \centering
        \includegraphics[width=\linewidth]{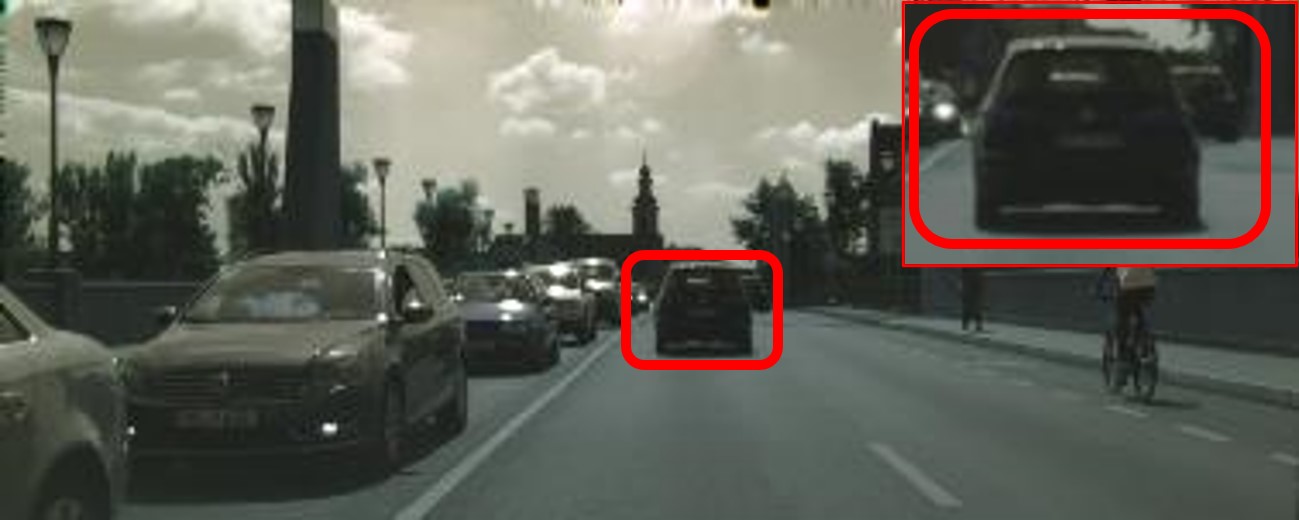}
        \subcaption{Input image}
        \label{fig:depth_max_rgb}
    \end{minipage}
    \begin{minipage}[t]{.47\linewidth}
        \centering
        \includegraphics[width=\linewidth]{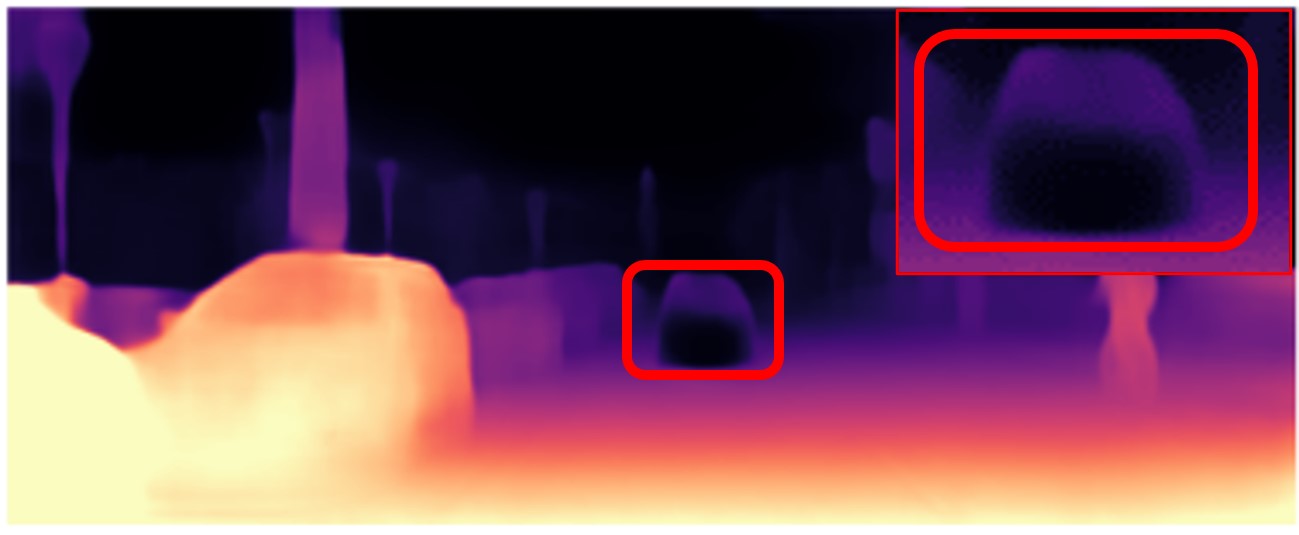}
        \subcaption{Depth estimation error}
        \label{fig:depth_max}
    \end{minipage}
    \caption{Estimation error by Depth Network.
    Even with a trained Depth Network,  sometimes estimates a miss value. 
    In this example, the maximum depth value for the car outlined in red is 58.61. 
    This value is notably high in the context of the depth distribution and can negatively affect the depth range settings.}
    \label{fig:overfit_depth}
\end{figure}

\begin{figure}[tp]
    \centering
    \hspace*{-0.4cm} 
    \includegraphics[width=8.3cm]{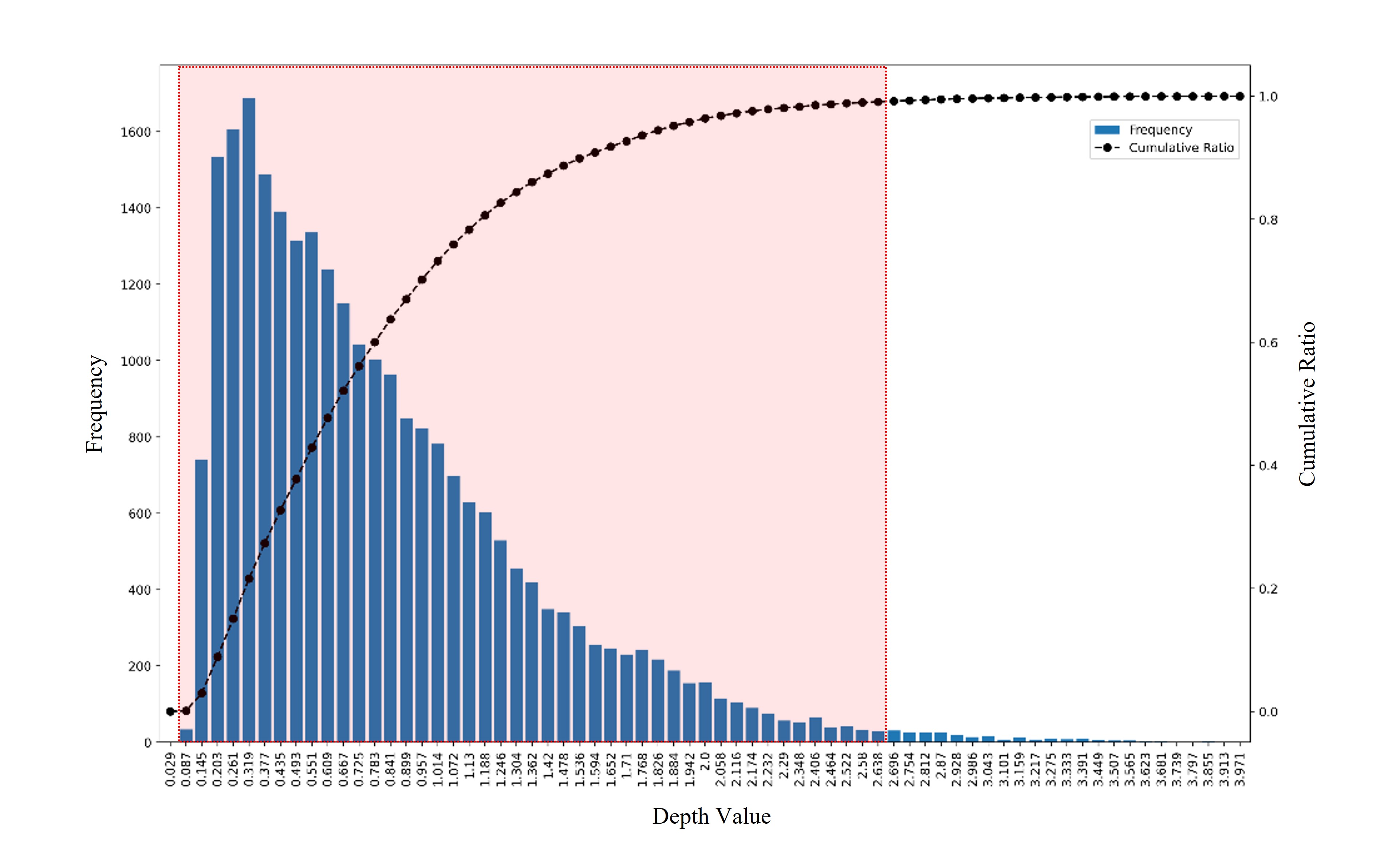}
    \caption{Distribution of the average depth values for each moving object in the VCAS-Motion Dataset.
    This graph visualizes the distribution of depth values within the range of 0.0 to 4.0. 
    The red region includes the depth range of $d_{min}=0.091$ to $d_{max}=2.646$ that was used in our experiments.}
    \label{fig:depth_analyse}
\end{figure}

In our work, we introduce modifications to the ManyDepth framework\cite{watson2021temporal}, which leverages outputs from single-image Depth Networks. 
Our key adaptation is tailored for motion segmentation, wherein the cost volume computation specifically hones in on the depth range containing moving objects. 
In ManyDepth, the values for $d_{min}$ and $d_{max}$ are determined as follows:
\begin{equation}
    \label{manydepth_dmax}
    d_{min} = 0.99 * d_{min} + 0.01 * \hat{d}^i_{min}
\end{equation}

\begin{equation}
    \label{manydepth_dmin}
    d_{max} = 0.99 * d_{max} + 0.01 * \hat{d}^i_{max}
\end{equation}

where $\hat{d}^i_{min}$ and $\hat{d}^i_{max}$ denote the average of the minimum and maximum depth values output by the Depth Network at iteration $i$, respectively.
However, as shown in \Cref{fig:overfit_depth}, the Depth Network occasionally produces output errors. 
Moreover, while \Cref{manydepth_dmax} and \Cref{manydepth_dmin} are applied across the entire image, they include depth values from the background regions. 
Yet, for constructing a Cost Volume for moving object detection, only the depth range where the moving object exists is crucial.
We therefore determined the depth range by considering only the regions where moving objects exist in the depth values output by the Depth Network and by excluding outliers.
\Cref{fig:depth_analyse} shows the depth value distribution for moving objects in the training data for the VCAS-Motion Dataset\cite{siam2021video}.
The Depth range set by the ManyDepth manner is [0.114, 17.95], whereas our method sets the depth range to [0.091, 2.646].
We set $d_{min}$ and $d_{max}$ based on the 1st and 99th percentiles of the depth distribution of moving objects in each dataset.
The depth value distribution is constructed by taking the average depth value within the ground truth mask region of the moving object as its depth value.
Notably, the learning of depth is conducted for the pre-training of the backbone and for determining the depth range, but the Depth Network is not utilized during the motion segmentation training. 

\subsection{Training Loss}
Following \cite{Cheng2022SparseInst}, the training loss is defined as:

\begin{equation}
L = L_M + \lambda_C L_C + \lambda_S L_S
\end{equation}

where $L_M$ is mask loss composed by dice loss\cite{milletari2016v} $L_D$ and pixel-wise binary cross entropy loss $L_P$:

\begin{equation}
L_M = \lambda_D L_D + \lambda_P L_P 
\end{equation}

$L_C$ is classification loss for estimated object class and uses focal loss\cite{lin2017focal} and $L_S$ denotes binary cross entropy loss for the objectness.
The $\lambda$ in each equation represents a coefficient, we set $\lambda_C = 2.0$, $\lambda_S = 3.0$, $\lambda_D = 2.0$, and $\lambda_P = 5.0$. 

\label{sec:method}

\section{EXPERIMENTS} \label{Experiments}

\subsection{Dataset and Evaluation Metrics} \label{data and metrics}
\noindent \textbf{KITTI} \cite{geiger2013vision} is a real-world dataset for autonomous driving.
This dataset is annotated with masks for moving cars in 200 images.
Following prior research, we report the F-measure\cite{dave2019towards} as an evaluation metric for instance segmentation masks of moving objects, and the IoU \cite{lv2018learning,ranjan2019competitive} for the non-moving background regions mask.
We utilize the pre-trained weights from the Depth Network of the ManyDepth framework for the Backbone.

\noindent \textbf{VCAS-Motion} \cite{siam2021video} is a dataset that has been annotated with motion segmentation labels for Cityscapes-VPS\cite{kim2020video} and KITTIMOTS\cite{voigtlaender2019mots}.
In this dataset, annotation masks are provided for moving objects within eight categories across 11,008 images.
Following previous works\cite{siam2021video}, we adopt class-agnostic quality as our evaluation metric, defined by the following equations: %\cite{siam2021video,wong2020identifying}

\begin{equation}
SQ =  \frac{\sum_{p,g \in TP} IoU(p, g)}{|TP|} \\
\end{equation}

\begin{equation}
RQ = \frac{|TP|}{|TP| + |FN|}
\end{equation}

\begin{equation}
CAQ = SQ \cdot RQ
\end{equation}

TP and FN represent the true positive and false negative, respectively.
An instance mask is deemed a true positive only if its intersection over union with the ground-truth exceeds 0.5.
We utilize the pre-trained weights from the Depth Network of the ManyDepth framework for the Backbone.

\subsection{Implementation Details} \label{Implementation details}

\begin{table*}[t]
  \centering
  \vspace{0.2cm} 
  \caption{
  Details for training hyperparameters.
  The gradient accumulation column indicates how many iterations are performed before updating the model.
  }
  \resizebox{2.\columnwidth}{!}{
  \begin{tabular}{lccccc}
  \toprule
  Training Session & Epochs & Learning Rate & Batch Size / GPU & Gradient Accumulation & Depth Range\\
  \midrule
  Depth and Pose (ManyDepth) & 20 & 1$\times 10^{-4}$ & 14 & - & -\\
  KITTI Dataset & 20 &  backbone: 5$\times 10^{-5}$, other: 1$\times 10^{-4}$ & 2 & 4 & [0.090, 2.465]\\
  VCAS-Motion 320$\times$960 & 15 & backbone: 5$\times 10^{-5}$, other: 1$\times 10^{-4}$ & 6 & 4 & [0.091, 2.646] \\
  VCAS-Motion KITTI  & 5 &  backbone: 1$\times 10^{-5}$, other: 5$\times 10^{-5}$ & 2 & 8 & [0.081, 2.424] \\
  VCAS-Motion Cityscapes & 5 &  backbone: 1$\times 10^{-5}$, other: 5$\times 10^{-5}$ & 2 & 8 & [0.101, 2.444] \\
  \bottomrule
  \end{tabular}

  }
  \vspace{-0.05in}
  \label{tab:training hyperparameters}
\end{table*}

We implemented our model using PyTorch\cite{paszke2017automatic} and trained it on two \underline{NVIDIA RTX A6000 GPUs}.
For inference speed, we measured frames per second (FPS) using a single \underline{NVIDIA RTX 3080Ti Mobile GPU} for all models used in our experiments.
All training employed AdamW as an optimizer, with $\beta_1$=0.9, $\beta_2$=0.999, weight decay of 0.001, and scheduled by cosine learning rate decay scheduler \cite{loshchilov2017sgdr}. 
The detailed hyperparameters for each training session are provided in \Cref{tab:training hyperparameters}.
For all training phases, we utilize data augmentation techniques, including horizontal flips and random variations in brightness, contrast, saturation, and hue. 
The respective ranges are $\pm0.2$ for brightness, contrast, and saturation, and $\pm0.1$ for hue.
We set the number of depth bins for Cost Volume construction at 64. 
We employed the ManyDepth framework for our Pose Network and trained it using all accessible data from KITTI and Cityscapes, or a combination of both, excluding their test sets. 
During this process, the depth distribution of moving objects from the Depth Network was utilized to define the depth range for the Cost Volume. 
It's important to note that the Depth Network was not utilized during the learning or inference phases for motion segmentation.

In the VCAS-Motion Dataset, Cityscapes is sampled every 5 frames, and we adopted a 2-frame interval for both training and evaluation to match the frame rate with KITTI.
Additionally, due to the resolution discrepancies between KITTI and Cityscapes, the size of the Cost Volume differ. 
To address this, both datasets were first trained jointly at a resolution of 320 $\times$ 960. 
Subsequent training was conducted on each dataset at its corresponding resolution.

\subsection{Results}

Following prior studies, we compare our proposed method with other models.
To show that our model is both efficient and effective in motion segmentation, we evaluate the FPS and the number of model parameters, 
in addition to the metrics presented in Section \ref{data and metrics}.

\begin{table}[t]
  \centering
  \caption{
  Results on the KITTI Dataset.
  Obj F represents the F-measure for the moving object masks. 
  Bg IoU denotes the Intersection over Union (IoU) for the background, which excludes the moving objects.
  Evaluations of models other than SparseInst and TMO are described in accordance with the assessments reported in \cite{yang2021learning}.
  TMO is a state-of-the-art model in the video object segmentation tasks, utilizing RGB images and optical flow as inputs to produce segmentation masks.
  For optical flow estimation, we utilize RAFT\cite{teed2020raft}.
  Similar to our proposed model, SparseInst has been adapted to process two consecutive frames as input.
  It dependently extracts features from each frame using the backbone, followed by concatenating these features before feeding them into the context encoder.
  }
  \resizebox{\columnwidth}{!}{%
  \begin{tabular}{l|c|c|c|c}
  \toprule
  Method & Params$\downarrow$ & FPS$\uparrow$ & Obj F $(\%) \uparrow$ & Bg IoU $(\%)\downarrow$ \\
  \midrule
  competitive collaboration \cite{ranjan2019competitive} & 5.22M & 196.08 & 50.87 & 85.50 \\
  COSNet \cite{lu2019see} & 81.24M & 4.09 & 66.67 & 93.03 \\
  MAT-Net \cite{zhou2020motion} & 142.69M & 3.75 & 68.40 & 93.08 \\
  FusionSeg \cite{dutt2017fusionseg} & 85.19M & 4.61 & 85.08 & 96.27 \\
  TMO \cite{cho2023treating} & 38.85M & 10.55 & 80.75 & 98.16 \\
  Rigidmask \cite{yang2021learning} & 138.52M & 4.92 & 90.71 & 97.05 \\
  SparseInst \cite{Cheng2022SparseInst} & 35.73M & 43.48 & 65.15 & 98.17 \\
  \midrule
  Ours & 35.71M & 19.12 & 84.62 & 98.40 \\
  \bottomrule
  \end{tabular}

  }
  \vspace{-0.05in}
  \label{tab:KITTI result}
\end{table}

\Cref{tab:KITTI result} shows the results for the KITTI dataset.
Compared to the state-of-the-art method, Rigidmask, our model achieves approximately four times the FPS and operates with roughly one-quarter of the parameters. 
Yet, the disparity in F-measure for moving objects is $6.09\%$. 
Additionally, our model exceeds Rigidmask in IoU for the background. 
We also observed that our model consistently outperforms other models in the balance between efficiency and precision.
Compared to the base model SparseInst, our model, while having a similar number of parameters, achieves $19.47\%$ improvement in F-measure.

\begin{table*}[t]
  \centering
  \vspace{0.2cm} 
  \caption{Results on the VCAS-Motion Dataset.
  We report the SQ, RQ, and CAQ metrics for KITTI and Cityscapes, respectively. 
  The FPS was measured on images with a resolution of 320$\times$960.
  The $+\alpha$ in VCANet indicates the inference time of Ego-Flow Suppression in terms of FPS.
  VCANet consists of the Motion Segmentation Network with 92.08M parameters and the network for Ego-flow Suppression estimation with 190.35M parameters. 
  The parameter breakdown for the Ego-flow Suppression estimation is as follows: FlowNet2\cite{ilg2017flownet} utilized for optical flow estimation has 162.51M, 
  MonoDepth2\cite{godard2019digging}'s Depth Network comprises 14.84M, and its Pose Network contains 13.00M, culminating in a total of 190.35M.
  The number of parameters in our model is 22.71M for Motion Segmentation Network and 13.00M for the Pose Network.}
  \resizebox{2.\columnwidth}{!}{%
  \begin{tabular}{l|c|c|ccc|ccc}
    \toprule
  & & & \multicolumn{3}{c}{KITTI} & \multicolumn{3}{|c}{Cityscapes} \\
  \cline{4-9}
  \multicolumn{1}{l|}{Method} & Params$\downarrow$ & FPS$\uparrow$ & SQ $(\%) \uparrow$ & RQ$(\%)\uparrow$ & CAQ$(\%)\uparrow$ & SQ $(\%) \uparrow$ & RQ$(\%)\uparrow$ & CAQ$(\%)\uparrow$ \\
  \midrule
  \multicolumn{1}{l|}{VCANet \cite{siam2021video}} & 92.08M+190.35M & 6.23$+\alpha$ & 82.4 & 75.4 & 62.1 & 78.0 & 56.2 & 43.8 \\
  \multicolumn{1}{l|}{SparseInst \cite{Cheng2022SparseInst}} & 35.73M & 47.62 & 79.96 & 35.21 & 28.15 & 68.52 & 21.34 & 14.62 \\
  \midrule
  \multicolumn{1}{l|}{Ours} & 22.71M + 13.00M & 26.89 & 79.44 & 55.24 & 43.88 & 77.85 & 55.65 & 43.32 \\
  \bottomrule
  \end{tabular}
  }
  \label{tab:VCAS-Motion result}
\end{table*}

In \Cref{tab:VCAS-Motion result}, we show results for the VCAS-Motion Dataset.
Compared to the state-of-the-art method, VCANet, our proposed model achieves over four times the FPS and less than one-quarter of the Motion Segmentation Network parameters, while maintaining nearly equivalent accuracy on Cityscapes.
Furthermore, considering the need for VCANet to estimate ego-flow suppression as an input, the difference is believed to be even more pronounced.
On the other hand, in the KITTI, our model underperforms compared to VCANet in CAQ by $18.22\%$.
Compared to the base model SparseInst, our model outperforms by $20.03\%$ in RQ and $15.73\%$ in CAQ on the KITTI dataset. 
Similarly, on the Cityscapes dataset, it outperforms by $34.31\%$ in RQ and $28.70\%$ in CAQ.

\subsection{Ablation Analysis} \label{Ablation}

\begin{figure}[tp]
  \centering
  \includegraphics[width=0.9\linewidth]{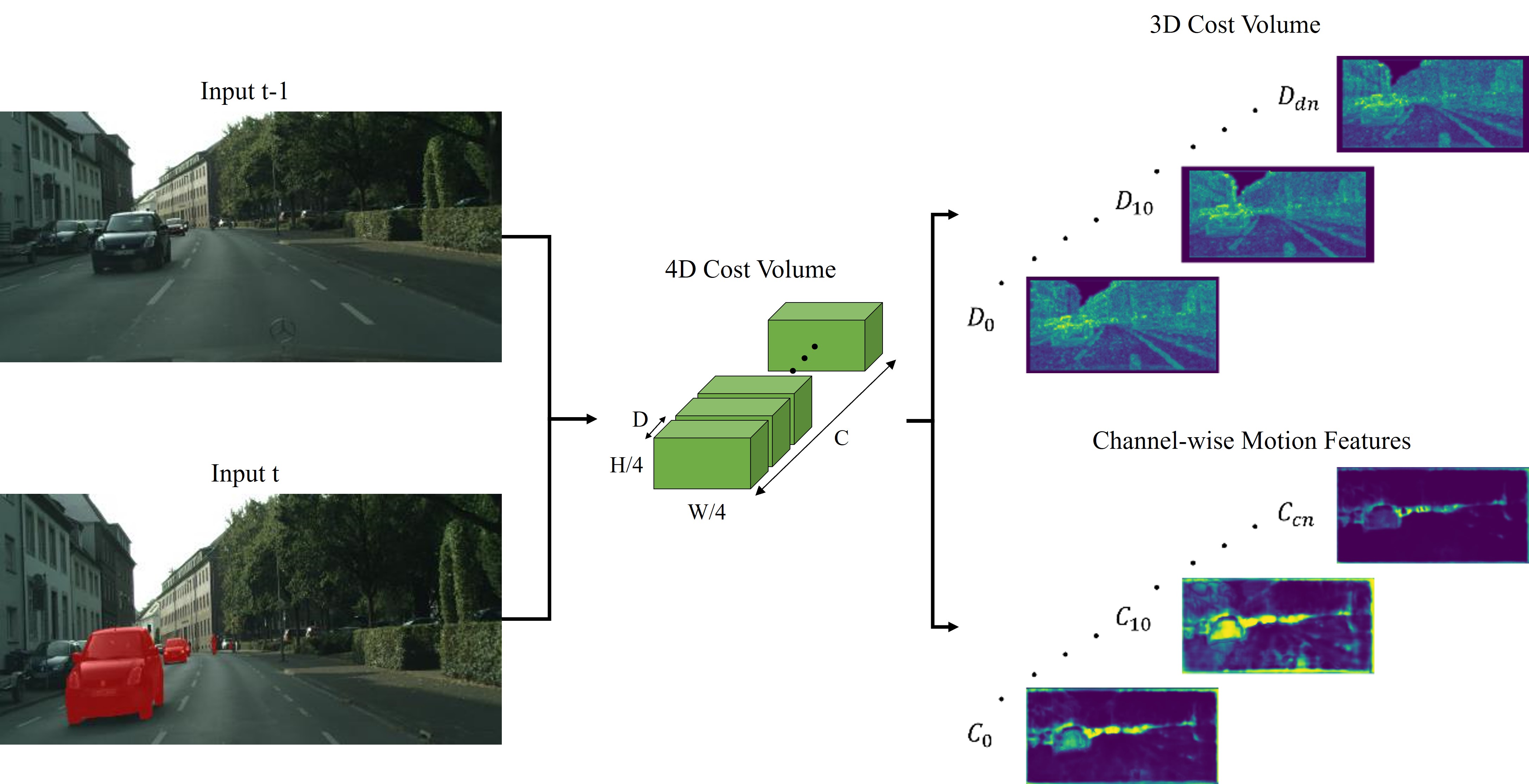}
  \caption{Visualization of Feature Maps for Channel-wise Motion Features and 3D Cost Volume.
  Channel-wise Motion Features shows activation for each moving instance, whereas 3D Cost Volume shows activation across the entire image. 
  In the input image t, the red mask indicates moving objects. 
  Channel-wise Motion Features here is represented as a feature map that has undergone 2D convolution without changing the dimension of the feature volume after the 3D Motion Extraction Network.}
  \label{fig:channel_and_3d}
\end{figure}

To demonstrate the effectiveness of Channel-wise Motion Features, we conduct two ablation analyses:

\noindent 1. Comparison of Channel-wise Motion Features and 3D Cost Volume. Channel-wise Motion Features not only captures the property of the 3D Cost Volume that takes the 
"inconsistency value of the moving object region" but also aggregates information in the depth direction for each channel. We verify its effects based on experiments.

\noindent 2. Comparison regarding depth range settings. We examine to what extent the depth range set for constructing the Cost Volume affects the accuracy of motion segmentation.

\noindent \textbf{Comparison between Channel-wise Motion Features and 3D Cost Volume.}

\begin{table}[t]
  \centering
  \caption{Comparison between using 3D Cost Volume and Channel-wise Motion Features on the VCAS-Motion Dataset.}
  \resizebox{\columnwidth}{!}{%
  \begin{tabular}{l|ccc|ccc}
    \toprule
    & \multicolumn{3}{c|}{KITTI} & \multicolumn{3}{c}{Cityscapes} \\
    \cline{2-7}
    Method & SQ $(\%) \uparrow$ & RQ$(\%)\uparrow$ & CAQ$(\%)\uparrow$ & SQ $(\%) \uparrow$ & RQ$(\%)\uparrow$ & CAQ$(\%)\uparrow$ \\
    \midrule
    3D Cost Volume & 73.08 & 49.70 & 36.32 & 76.08 & 49.13 & 37.38 \\
    %\midrule
    Our setting methods & 79.44 & 55.24 & 43.88 & 77.85 & 55.65 & 43.32 \\
    \bottomrule
  \end{tabular}
  }
  \label{tab:3dcost_comparison}
\end{table}

In this experiment, the 3D Cost Volume is constructed by taking the average of the channel direction of the 4D Cost Volume, the same as ManyDepth.

\Cref{fig:channel_and_3d} illustrates the differences in the feature maps of Channel-wise Motion Features and the 3D Cost Volume.  
Channel-wise Motion Features aggregates information in the Depth direction for each channel of the feature maps between consecutive frames, and it exhibits activations to moving objects.
This indicates that it is capable of capturing information regarding the motion of each instance.
Furthermore, as shown in \Cref{tab:3dcost_comparison}, on the VCAS-Motion Dataset, the use of Channel-wise Motion Features yields better results compared to the use of 3D Cost Volume.

\noindent \textbf{Effects of setting $d_{min}$ and $d_{max}$ for motion segmentation.}

We compare the depth range setting method we proposed with the method proposed in ManyDepth. 
In \Cref{tab:depth_range_comparison}, we show the results obtained when applying both depth range settings to our model. 
Our method outperforms the setting approach of ManyDepth across all evaluation metrics, indicating that limiting the range setting to the moving object detection area is effective for motion segmentation.

\begin{table}[t]
  \centering
  \caption{Comparison of depth range settings.
  In the ManyDepth approach, the depth range for VCAS-Motion $320 \times 960$ is [0.114, 17.95], that for KITTI is [0.105, 8.471], and that for Cityscapes is [0.105, 27.05]. 
  Our settings are the same as in \Cref{tab:training hyperparameters}. }
  \resizebox{\columnwidth}{!}{
  \begin{tabular}{l|ccc|ccc}
    \toprule
    & \multicolumn{3}{c|}{KITTI} & \multicolumn{3}{c}{Cityscapes} \\
    \cline{2-7}
    Depth Range & SQ $(\%) \uparrow$ & RQ$(\%)\uparrow$ & CAQ$(\%)\uparrow$ & SQ $(\%) \uparrow$ & RQ$(\%)\uparrow$ & CAQ$(\%)\uparrow$ \\
    \midrule
    ManyDepth manner & 77.89 & 49.86 & 38.84 & 77.33 & 54.53 & 42.17 \\
    Our setting methods & 79.44 & 55.24 & 43.88 & 77.85 & 55.65 & 43.32 \\
    \bottomrule
  \end{tabular}
  }
  \label{tab:depth_range_comparison}
\end{table}

\label{sec:experiment}

\section{CONCLUSION}
We proposed Channel-wise Motion Features, a novel motion feature representation for motion segmentation. 
We tackled the previously unattained trade-off between inference speed and accuracy, demonstrating the efficiency and effectiveness of our proposed model. 
Experiments showed that our model could achieve approximately four times the FPS and reduce parameters by about 25$\%$, with only a 6.09$\%$ F-measure drop on the KITTI Dataset 
and a 0.48$\%$ CAQ decrease on the VCAS-Motion Cityscapes Dataset compared to state-of-the-art models.
Furthermore, through ablation analysis, we showed that Channel-wise Motion Features effectively captures features of moving objects. 
We expect that our findings will lead to advancements in motion segmentation and applications such as autonomous driving.

\noindent \textbf{Limitations.}
Similar to previous methods based on instance activation \cite{Cheng2022SparseInst,he2023fastinst}, our model exhibits a weakness in detecting smaller objects. 
Additionally, distant moving objects, which exhibit subtle apparent motions, are not detected. 
Potential solutions include employing high-resolution feature maps when constructing the Cost Volume. 
However, such approaches significantly increase the computational cost.
Additionally, since processing the 4D Cost Volume with 3D convolutions requires a large amount of GPU memory, it is necessary to explore practical solutions to address this issue. %%%%%%%%%%%%%%%%%%%%%
We intend to address these challenges in our future research.

\label{sec:conclusion}

\bibliographystyle{IEEEtran}
\bibliography{11_references}
\addtolength{\textheight}{-12cm}  
\end{document}